%% file: ijcai22.tex
\documentclass{article}
\usepackage{ijcai22}

\input{math_commands.tex}

\usepackage{times}
\usepackage{soul}
\usepackage{url}
\usepackage[colorlinks, linkcolor=red, anchorcolor=red, citecolor=cyan, breaklinks]{hyperref}
\usepackage[utf8]{inputenc}
\usepackage[small]{caption}
\usepackage[table]{xcolor}

\usepackage{graphicx}
\usepackage{amsmath}
\usepackage{amsthm}
\usepackage{booktabs}
\usepackage{mydef}
\usepackage{subfiles}
\usepackage{float}
\usepackage{stfloats}
\usepackage{wrapfig}
\newcommand{\tabincell}[2]{\begin{tabular}{@{}#1@{}}#2\end{tabular}}

\title{A Survey on Protein Representation Learning: Retrospect and Prospect}

\author{
    Lirong Wu {$^{1,2}$ \thanks{Equal contribution, $^{\dagger}$ Corresponding author}} \and Yufei Huang {$^{1,2\ *}$ } \and Haitao Lin {$^{1,2}$ } \and Stan Z. Li {$^{1 \dagger}$} 
    \affiliations
    {$^{1}$} AI Lab, Research Center for Industries of the Future, Westlake University \\
    {$^{2}$} College of Computer Science and Technology, Zhejiang University \\
    \emails
    \small\texttt{\{wulirong,huangyufei,linhaitao,stan.zq.li\}@westlake.edu.cn}
}

\begin{document}

\maketitle

\begin{abstract}
    Proteins are fundamental biological entities that play a key role in life activities. The amino acid sequences of proteins can be folded into stable 3D structures in the real physicochemical world, forming a special kind of \emph {sequence-structure data}. With the development of Artificial Intelligence (AI) techniques, \emph{Protein Representation Learning} (PRL) has recently emerged as a promising research topic for extracting informative knowledge from massive protein sequences or structures. To pave the way for AI researchers with little bioinformatics background, we present a timely and comprehensive review of PRL formulations and existing PRL methods from the perspective of model architectures, pretext tasks, and downstream applications. We first briefly introduce the motivations for protein representation learning and formulate it in a general and unified framework. Next, we divide existing PRL methods into three main categories: sequence-based, structure-based, and sequence-structure co-modeling. Finally, we discuss some technical challenges and potential directions for improving protein representation learning. The latest advances in PRL methods are summarized in a GitHub repository \url{https://github.com/LirongWu/awesome-protein-representation-learning}.
\end{abstract}

\section{Introduction}
Proteins perform specific biological functions that are essential for all living organisms and therefore play a key role when investigating the most fundamental questions in the life sciences. The proteins are composed of one or several chains of amino acids that fold into a stable 3D structure to enable various biological functionalities. Therefore, understanding, predicting, and designing proteins for biological processes are critical for medical, pharmaceutical, and genetic research.

Previous approaches on protein modeling are mostly driven by biological or physical priors, and they explore complex sequence-structure-function relationships through energy minimization \cite{rohl2004protein,xu2011improving}, dynamics simulations \cite{hospital2015molecular,karplus1990molecular}, etc. With the development of artificial intelligence and low-cost sequencing technologies, data-driven \emph{Protein Representation Learning} (PRL) \cite{jumper2021highly,rao2019evaluating,rives2021biological,hermosilla2022contrastive,jing2020learning} has made remarkable progress due to its superior performance in modeling complex nonlinear relationships. The primary goal of protein representation learning is to extract transferable knowledge from protein data with well-designed model architectures and pretext tasks, and then generalize the learned knowledge to various protein-related downstream applications, ranging from structure prediction to sequence design. Despite their great progress, it is still tricky for AI researchers without bioinformatics background to get started with protein representation learning, and one obstacle is the vast amount of physicochemical knowledge involved behind the proteins. Therefore, a survey on PRL methods that is friendly to the AI community is urgently needed.

Existing surveys related to PRL \cite{iuchi2021representation,unsal2020evaluation,hu2021survey,torrisi2020deep} are mainly developed from the perspective of biological applications, but do not go deeper into other important aspects, such as model architectures and pretext tasks. Overall, our contributions can be summarized as follows: \textbf{(1)} \emph{Comprehensive review.} Our survey provides a comprehensive and up-to-date review of existing PRL methods from the perspective of the model architectures and pretext tasks. \textbf{(2)} \emph{New taxonomy.}  We divide existing PRL methods into three categories: sequence-based, structure-based, and sequence-structure co-modeling. \textbf{(3)} \emph{Detailed Implementations.} We summarize the paper lists and open-source codes in a public GitHub repository, setting the stage for the development of more future works. \textbf{(4)} \emph{Future directions.} We point out the technical limitations of current research and discuss several promising directions.

\section{Notation and Problem Statement}
The sequence of amino acids can be folded into a stable 3D structure, forming a special kind of \emph {sequence-structure data}, which determines its properties and functions. Therefore, we can model each protein as a graph $\mathcal{G} = (\mathcal{V}, \mathcal{E}, \mathcal{X}, \mathcal{F})$, where $\mathcal{V}$ is the \emph{ordered set} of $N$ nodes in the graph representing amino acid residues and $\mathcal{E} \in \mathcal{V} \times \mathcal{V}$ is the set of edges that connects the nodes. Each node $u\in\mathcal{V}$ in graph $\mathcal{G}$ can be attributed with a scalar-vector tuple $\mathbf{x}_u=(s_u, V_u)$, where $s_u \in \mathbb{R}^O$ and $V_u \in \mathbb{R}^{3\times P}$. Each edge $e \in \mathcal{E}$ can be attributed with a scalar-vector tuple $\mathbf{f}_e=(s_e, V_e)$, where $s_e \in \mathbb{R}^T$ and $V_e \in \mathbb{R}^{3\times D}$. 

\begin{figure}[!htbp]
	\begin{center}
		\includegraphics[width=1.0\linewidth]{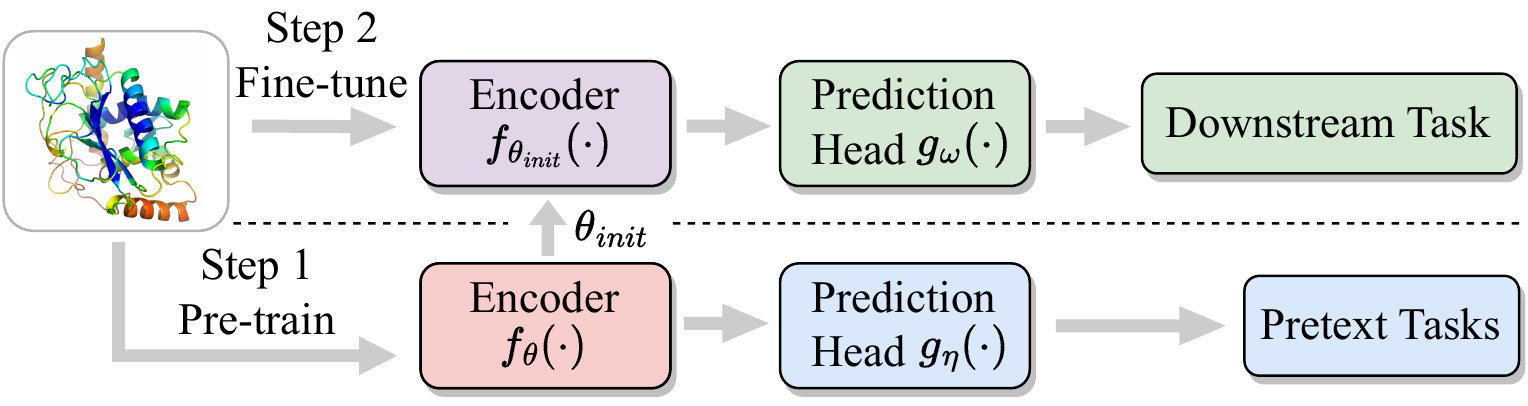}
	\end{center}
	\caption{A general framework for protein representation learning.}
	\label{fig:1}
\end{figure}

Given a model architecture $f_{\theta}(\cdot)$ and a set of $K$ losses of pretext tasks $\{\mathcal{L}_{pre}^{(1)}(\theta,\eta_1),\mathcal{L}_{pre}^{(2)}(\theta,\eta_2),\cdots,\mathcal{L}_{pre}^{(K)}(\theta,\eta_K)\}$ with projection heads $\{g_{\eta_k}(\cdot)\}_{k=1}^K$, \emph{Protein Representation Learning} (PRL) usually works in a two-stage manner: (1) Pre-training the model $f_\theta(\cdot)$ with pretext tasks; and (2) Fine-tuning the pre-trained model $f_{\theta_{init}}(\cdot)$ with a projection head $g_\omega(\cdot)$ under the supervision of a specific downstream task $\mathcal{L}_{task}(\theta,\omega)$. The learning objective can be formulated as
\begin{equation}
\begin{aligned}
\theta^{*}, \omega^{*}= & \arg \min _{(\theta, \omega)} \mathcal{L}_{task}(\theta_{init}, \omega), \\ \text{s.t.} \text{ } \text{ } \theta_{init}, \{\eta_k^*\}_{k=1}^K= &\mathop{\arg\min}_{\theta, \{\eta_k\}_{k=1}^K} \sum_{k=1}^K \lambda_k \mathcal{L}_{pre}^{(k)}(\theta,\eta_k)
\end{aligned}
\label{equ:1}
\end{equation}
where $\{\lambda_k\}_{k=1}^K$ are trade-off task hyperparameters. A high-level overview of the PRL framework is shown in Fig.~\ref{fig:1}. In practice, if we set $K=1$, $\omega\!=\!\eta_1$, i.e., $\mathcal{L}_{pre}^{(1)}(\theta,\eta_1)\!=\!\mathcal{L}_{task}(\theta,\omega)$, it is equivalent to learning task-specific representations directly under downstream supervision, which in this survey can be considered as a special case of Eq.~(\ref{equ:1}).

In this survey, we mainly focus on the model architecture $f_{\theta}(\cdot)$ and pretext tasks $\{\mathcal{L}_{pre}^{(k)}(\theta,\eta_k)\}_{k=1}^K$ for protein representation learning, and defer the discussion on downstream applications until Sec.~\ref{sec:6}. A high-level overview of this survey with some representative examples is shown in Fig.~\ref{fig:2}.

\subfile{overview.tex}

\section{Model Architectures}
In this section, we summarize some commonly used model architectures for learning protein sequences or structures.

\subsection{Sequence-based Encoder}
The sequence encoder takes as input $(\mathcal{V}, \mathcal{X})$ and then aims to capture the dependencies between amino acids. \cite{wang2019high} treats protein sequences as a special ``biological language" and then establishes an analogy between such ``biological language" and natural (textual) language. Inspired by this, many classical model architectures developed for natural language processing can be directly extended to handle protein sequences \cite{asgari2019deepprime2sec}. Depending on whether a single sequence or multiple sequences are to be encoded, there are a variety of different sequence-based encoders.

\subsubsection{Single Sequences}
The commonly used sequence encoders for modeling \textbf{\emph{single sequences}} include Variational Auto-Encoder (VAE) \cite{sinai2017variational,ding2019deciphering}, Recurrent Neural Networks (RNNs) \cite{armenteros2020language}, Long Short-Term Memory (LSTM) \cite{hochreiter1997long}, BERT \cite{devlin2018bert}, Transformer \cite{vaswani2017attention}. Based on the vanilla Transformer, \cite{wu2022high} proposes a novel geometry-inspired transformer (Geoformer) to further distill the structural and physical pairwise relationships between amino acids into the learned protein representation. If we do not consider the ordering of amino acids in the sequences, we can also directly apply Convolutional Neural Networks (CNNs) \cite{lecun1995convolutional} or ResNet \cite{he2016deep} to capture the local dependencies between adjacent amino acids.

\subsubsection{MSA Sequences}
The long-standing practices in computational biology are to make inferences from a family of evolutionarily related sequences \cite{weigt2009identification,thomas2005graphical,lapedes1999correlated}. Therefore, there have been several \textbf{\emph{multiple sequences}} encoders proposed to capture co-evolutionary information by taking as input a set of sequences in the form of multiple sequence alignment (MSA). For example, MSA Transformer \cite{rao2021msa} extends the self-attention mechanism to the MSA setting, which interleaves self-attention across rows and columns to capture dependencies between amino acids and between sequences. As a crucial component of AlphaFold2, Evoformer \cite{jumper2021highly} alternatively updates \textit{MSA} and \textit{Pair} representations in each block, which encode co-evolutionary information in sequences and relations between residues, respectively.

\subsection{Structure-based Encoder}
Despite the effectiveness of sequence-based encoders, the power of pre-training with protein structures has been rarely explored, even though protein structures are known to be determinants of protein functions. To better utilize this critical structural information, a large number of structure-based encoders have been proposed to model structural information, which can be mainly divided into three categories: feature map-based, message-passing GNNs, and geometric GNNs.

\subsubsection{Feature map-based Methods}
The use of deep learning to model protein 3D structures could be traced back to a decade ago \cite{zhang2010novel,schaap2001rosetta}. Early methods directly extracted several hand-crafted \textbf{\emph{feature maps}} from protein structures and then applied 3D CNNs to model the geometric information of proteins \cite{derevyanko2018deep,amidi2018enzynet,townshend2019end}. Later work extended 3D CNNs to spherical convolution for identifying interaction patterns on protein surfaces \cite{sverrisson2021fast,gainza2020deciphering}.

\subsubsection{Message-passing GNNs}
To further capture the geometric relationships and biomedical interactions between amino acids, it has been proposed to first construct a graph from the extracted feature maps by thresholding or $k$ Nearest Neighbors ($k$NN) \cite{preparata2012computational}. Then, many existing message-passing Graph Neural Networks (GNNs) can be directly applied to model protein structures, including Graph Convolutional Network (GCN) \cite{kipf2016semi}, Graph Isomorphism Network (GIN) \cite{xu2018powerful}, and GraphSAGE \cite{hamilton2017inductive}. However, the edges in the protein graph may have some key properties, such as dihedral angles and directions, which determine the biological function of proteins. With this in mind, there have been several structure-based encoders proposed to simultaneously leverages the node and edge features of the protein graph. For example, \cite{hermosilla2020intrinsic} proposes IE convolution (IEconv) to simultaneously capture the primary, secondary and tertiary structures of proteins by incorporating intrinsic and extrinsic distances between nodes. Besides, \cite{hermosilla2022contrastive} adopts a similar architecture to IEConv, but introduces seven additional edge features to efficiently describe the relative position and orientation of neighboring nodes. Furthermore, GearNet \cite{zhang2022protein} proposes a simple structure encoder, which encodes spatial information by adding different types of sequential or structural edges and then performs both node-level and edge-level message passing simultaneously.

\subsubsection{Geometric GNNs}
The above message-passing GNNs incorporate the 3D geometry of proteins by encoding the vector features $V_u$/$V_e$ into rotation-invariant scalars $s_u$/$s_e$. However, reducing this vector information directly to scalars may not fully capture complex geometry. Therefore, geometric-aware neural networks are proposed to bake 3D rigid transformations into network operations, leading to SO(3)-invariant and equivariant GNNs. For example, \cite{jing2020learning} introduces Geometric Vector Perceptrons (GVPs), which replace standard multi-layer perceptrons (MLPs) in feed-forward layers and operate directly on both scalar and vector features under a global coordinate system. Besides, \cite{aykent2022gbpnet} proposes Geometric Bottleneck
Perceptron (GBPs) to integrate geometric features and capture complex geometric relations in the 3D structure, based on which a new SO(3)-equivariant message passing neural network is proposed to support a variety of geometric representation learning tasks. To achieve more sensitive geometric awareness in both global transformations and local relations, \cite{li2022directed} proposes Directed Weight Perceptrons (DWPs) by extending not only the hidden neurons but the weights from scalars to 2D/3D vectors, naturally saturating the network with 3D structures in the Euclidean space.

\subsection{Sequence-structure Encoder}
Compared to sequence- and structure-based encoders, comparatively less work has focused on the co-encoding of protein sequences and structures. The mainstream model architecture is to extract amino acid representations as node features by a language model and then capture the dependencies between amino acids using a GNN module. For example, \cite{gligorijevic2021structure} introduces DeepFRI, a Graph Convolutional Network (GCN) for predicting protein functions by leveraging sequence representations extracted from a protein language model (LSTM) and protein structures. Besides, LM-GVP \cite{wang2021lm} is composed of a protein language model (composed of Transformer blocks) and a GVP network, where the protein LM takes protein sequences as input to compute amino acid embeddings and the GVP network is used to make predictions about protein properties on a graph derived from the protein 3D structure. Moreover, \cite{you2022cross} applies the hierarchical RNN and GAT to encode both protein sequences and structures and proposes a cross-interaction module to enforce a learned relationship between the encoded embeddings of the two protein modalities.

\section{Pretext Task}
The pretext tasks are designed to extract meaningful representations from massive data through optimizing some well-designed objective functions. In this section, we summarize some commonly used pretext tasks for learning on proteins.

\subsection{Sequence-based Pretext Task}
There have been many pretext tasks proposed for pre-training language models, including Masked Language Modeling (MLM) and Next Sentence Prediction (NSP) \cite{devlin2018bert}, which can be naturally extended to pre-train protein sequences. We divide existing sequence-based pretext tasks into two main categories: self-supervised and supervised. 

\subsubsection{Self-supervised Pretext Task}
The self-supervised pretext tasks utilize the training data itself as supervision signals without the need for additional annotations. If we consider an amino acid in a sequence as a word in a sentence, we can naturally extend masked language modeling to protein sequences. For example, we can statically or dynamically mask out a single or a set of contiguous amino acids and then predict the masked amino acids from the remaining sequences \cite{rao2019evaluating,elnaggar2020prottrans,rives2021biological,rao2021msa,nambiar2020transforming,xiao2021modeling}. Besides, \cite{mcdermott2021adversarial} combines adversarial training with MLM and proposes to mask amino acids in a learnable manner. Taking into account the dependence between masked amino acids, Pairwise MLM (PMLM) \cite{he2021pre} proposes to model the probability of a pair of masked amino acids instead of predicting the probability of a single amino acid. Besides, Next Amino acid Prediction (NAP) \cite{alley2019unified,elnaggar2020prottrans,strodthoff2020udsmprot} aims to predict the type of the next amino acid based on a set of given sequence fragments. Different from the above methods, Contrastive Predictive Coding (CPC) \cite{lu2020self} applies different augmentation transformations on the input sequence to generate different views, and then maximizes the agreement of two jointly sampled pairs against that of two independently sampled pairs.

\subsubsection{Supervised Pretext Task}
The supervised pretext tasks use additional labels as auxiliary information to guide the model to learn knowledge relevant to downstream tasks. For example, PLUS \cite{min2021pre} devises a protein-specific pretext task, namely Same-Family Prediction (SFP), which trains a model to predict whether a given protein pair belongs to the same protein family. The protein family labels provide weak structural information and help the model learn structurally contextualized representations. Besides, \cite{sturmfels2020profile} proposes to use HMM profiles derived from MSA as labels and then take Profile Prediction as a pretext task to help the model learn information about protein structures. In addition, to leverage the exponentially growing protein sequences that lack costly structural annotations, Progen \cite{madani2020progen} trains a language model with conditioning tags that encode various annotations, such as taxonomic, functional, and locational information.

\subsection{Structure-based Pretext Task}
Despite the great progress in the design of structure-based encoders and graph-based pretext tasks \cite{wu2021self,xie2022self,liu2022graph}, there are few efforts focusing on the structure-based pre-training of proteins. Existing structure-based pretext tasks for proteins can be mainly classified into two branches: contrastive and predictive methods.

\subsubsection{Contrastive Pretext Task}
The primary goal of contrastive methods is to maximize the agreement of two jointly sampled positive pairs. For example, Multiview Contrast \cite{hermosilla2022contrastive} proposes to randomly sample two sub-structures from each protein, encoder them into two representations, and finally maximize the similarity between representations from the same protein while minimizing the similarity between representations from different proteins. Besides, \cite{zhang2022protein} adopts almost the same architecture as Multiview Contrast, but replaces GearNet with IEConv as the structure encoder.

\subsubsection{Predictive Pretext Task}
The contrastive methods deal with the \emph{inter-data} information (data-data pairs). In contrast, the predictive methods aim to self-generate informative labels from the data as supervision and handle the \emph{data-label} relationships. Categorized by different types of pseudo labels, the predictive methods have different designs that can capture different levels of structural protein information. For example, \cite{chen2022structure} proposes two predictive tasks, namely \emph{Distance Prediction} and \emph{Angle Prediction}, which take hidden representations of residues as input and aim to predict the relative distance between pairwise residues and the angle between two edges, respectively, which helps to learn structure-aware protein representations. Furthermore, \cite{hermosilla2022contrastive} propose \emph{Residue Type Prediction} and \emph{Dihedral Prediction} based on geometric or biochemical properties. Specifically, \emph{Residue Type Prediction} randomly masks the node features of some residues and then lets the structure-based encoders predict these masked residue types. Instead, \emph{Dihedral Prediction} constructs a learning objective by predicting the dihedral angle between three consecutive edges. Besides, \cite{you2022cross} proposes graph completion (GraphComp), which takes as input a protein graph with partially masked residues and then makes predictions for those masked tokens. 

\renewcommand{\dblfloatpagefraction}{.95}
\subfile{table_1.tex}

\subsection{Sequence-structure Pretext Task}
Most of the existing methods design pretext tasks for a single modality but ignore the dependencies between sequences and structures. If we can design the pretext task based on both protein sequences and structures, it should capture richer information than using single modality data. In practice, there is no clear boundary between pretext tasks and downstream tasks. For example, AlphaFold2 \cite{jumper2021highly} takes full-atomic structure prediction as a downstream task. However, if we are concerned with protein property prediction, structure prediction can also be considered as a pretext task that enables the learned sequence representations to contain sufficient structural information. It was found by \cite{hu2022exploring} that the representations from AlphFold2's Evoformer could work well on various protein-related downstream tasks, including fold classification, stability prediction, etc. Moreover, \cite{yang2022masked} proposes a novel pre-training pretext task, namely Masked Inverse Folding (MIF), which trains a model to reconstruct the original amino acids conditioned on the corrupted sequence and the backbone structure.

\section{Downstream Tasks (Applications)} \label{sec:6}
In the above, we have presented a variety of commonly used model architectures and pretext tasks for protein representation learning, based on which we summarized the surveyed works in Table.~\ref{tab:1}, listing their categories, model architectures, pretext tasks, and publication years. In this section, we can divide existing downstream tasks for protein representation learning into the following four main categories: protein property prediction, protein (complex) structure prediction, protein design, and structure-based drug design. 

It is worth noting that some downstream tasks have labels (i.e., model outputs) that do not change with rigid body transformations of the inputs (if they can, e.g., protein structures). For example, various protein property prediction tasks take a transformable protein structure as input and output a constant prediction, usually modeled as a simple multi-label classification problem or multiple binary classification problem. However, the labels of some downstream tasks will change equivariantly with the inputs, and these tasks are getting more and more attention. Typically, the learning objectives of these tasks are structure-related, and they usually have higher requirements on the model architecture, requiring the model to be SE(3)-equivariant. We believe that from the perspective of protein representation learning, the approaches to different downstream tasks can also learn from each other.

\subsection{Protein Property Prediction} The protein property prediction aims to regress or classify some important properties from protein sequences or structures that are closely related to biological functions, such as the types of secondary structure, the strength of connections between amino acids, types of protein folding, fluorescence intensity, protein stability, etc. \cite{rao2019evaluating}. Besides, several protein-specific prediction tasks can also be grouped into this category, including quality evaluation of protein folding \cite{baldassarre2021graphqa}, predicting the effect of mutations on protein function \cite{meier2021language}, and predicting protein-protein interactions \cite{wang2019high}.

\subsection{Protein (Complex) Structure Prediction} The primary goal of protein structure prediction is to predict the structural coordinates from a given set of amino acid sequences. Some approaches aim to predict only backbone coordinates \cite{baek2021accurate,si2020deep}, while others focus on the more challenging full-atomic coordinate predictions \cite{jumper2021highly,wu2022high,rao2021msa}. On the other hand, protein structure refinement \cite{hiranuma2021improved,wu2022atomic} proposes to update a coarse protein structure to generate a more fine-grained structure in an iterative manner. Besides, the task of protein structure inpainting aims to reconstruct the complete protein structure from a partially given sub-structure \cite{mcpartlon2022attnpacker} or distance map \cite{lee2022proteinsgm}.

\subsection{Protein Design} Deep learning-based protein design has made tremendous progress in recent years, and the major works can be divided into three categories. The first one is to pre-train the model with a large number of sequences from the same protein family, and then use it to generate new homologous sequences \cite{smith1990automatic}. The structure-based methods aim to directly generate the protein sequences under the condition of a given protein structure \cite{ingraham2019generative}. The last and most challenging one is the de novo protein design \cite{huang2016coming,korendovych2020novo,koepnick2019novo}, which aims to generate both protein sequences and structures conditioned on taxonomic and keyword tags such as molecular function and cellular component.

\subsection{Structure-Based Drug Design} Structure-Based Drug Design (SBDD) is a promising direction for fast and cost-efficient compound discovery. Specifically, SBDD designs inhibitors or activators (usually small molecules, i.e., drugs) directly against protein targets of interest, which means a high success rate and efficiency \cite{SBDDScience1992,SBDDScience2000}. In the past two years, a line of auto-regressive methods have been proposed for SBDD \cite{liu2022graphbp,peng2022pocket2mol,masuda2020generating}, which generate molecule atoms one by one conditioned on given structure context of protein targets. Recently, there are some works based on Denoising Diffusion Probabilistic Model (DDPM) \cite{lin2022diffbp,schneuing2022structure}. Targeting on specific protein pockets, the diffusion-based methods generate molecule atoms as a whole from random gaussian noise.

The above methods are all dependent on a proper representation module of protein, especially the protein structure. The early attempt of deep generative models in this field \cite{luo20213d} uses 3D CNN as the protein structure context encoder to get meaningful and roto-translation invariant features. With the development of protein structure representation methods, particularly the geometric-aware models, subsequent methods widely use geometric-(equi/in)variant networks, such as EGNN \cite{gong2019exploiting}, GVP \cite{jing2020learning}, and IPA \cite{jumper2021highly}, as the backbones. It is worth noting that protein representation models are not only common in various protein structure context encoders, but many generative decoders can also adopt its architectural design. From this example, we can see that protein representation is a very fundamental problem and that many downstream tasks involving proteins can benefit from advances of protein representation research in various aspects, including better embeddings and more excellent model architectures.

\section{Deep Insights and Future Outlooks}
\subsection{Deeper Insights}
On the basis of a detailed review of the model architectures, pretext tasks, and downstream tasks, we would like to provide some deeper insights into protein representation learning.

\subsubsection{Insights 1: PRL is the core of deep protein modeling}
With the development of deep learning, deep protein modeling is becoming a popular research topic, and one of its core is how to learn ``meaningful" representations for proteins. This involves three key issues: (1) Feature Extraction: model architectures; (2) Pre-training: pretext tasks; and (3) Application: downstream tasks. An in-depth investigation of the above three key issues is of great importance for the development of more deep protein modeling methods.

\subsubsection{Insights 2: Task-level convertibility}
Throughout this survey, one of the main points we have emphasized is the convertibility between downstream tasks and pretext tasks. We believe we are the first to explain the role of pretext tasks from this perspective, which seems to have been rarely involved in previous work. For example, we directly categorize some well-known downstream tasks, such as full-atomic structure prediction, as a specific kinds of pretext tasks. The motivation behind such an understanding lies in the fact that the definition of a task is itself a relative concept and that different tasks can help the model extract different aspects of information, which may be complementary to each other. For example, full-atomic structure prediction helps the model capture rich structural information, which is also beneficial for various protein property prediction tasks, such as folding prediction, since it is known that protein structure often determines protein function. This suggests that whether a specific task is a downstream task or a pretext task usually depends on what we are concerned about, and the role of a task may keep changing from application to application.

\subsubsection{Insights 3: Data-specific criterion for design selections}
It is tricky to discuss the advantages and disadvantages of different methods or designs because the effectiveness of different methods depends heavily on the size, format, and complexity of the data. For example, for simple small-scale data, Transformer is not necessarily more effective than traditional LSTM for sequence modeling, and the situation may be completely opposite for large-scale complex data. Therefore, there is no ``optimal" architecture or pretext task that works for all data types and downstream tasks, and the criterion for the selection of architecture and pretext task is data-specific.

\subsection{Future Outlooks}
Despite the great progress of existing methods, challenges still exist due to the complexity of proteins. In this section, we suggest some promising directions for future work.

\subsubsection{Direction 1: Broader application scenarios}
The biological research topics on proteins are diverse, but most of the existing work has delved into only a small subset of them, due to the fact that these topics have been well formalized by some representative works, such as AlphaFlod2 \cite{jumper2021highly} for protein structure prediction and TAPE \cite{rao2019evaluating} for protein property prediction. As a result, it is more worthwhile to explore the role of protein representation learning in a wider range of biological application scenarios than to design some overly complex modules for subtle performance gains in a well-formalized application.

\subsubsection{Direction 2: Unified evaluation protocols}
Research in protein representation learning is now in an era of barbarism. While a great deal of new works are emerging every day, most of them are on unfair comparisons, such as with different datasets, architectures, metrics, etc. For example, some MSA-based works on structure prediction have been blatantly compared with those single-sequence-based works and claimed to be better. To promote the health of the field, there is an urgent need to establish unified evaluation protocols in various downstream tasks to provide fair comparisons.

\subsubsection{Direction 3: Protein-specific designs}
Previous PRL methods directly take mature architectures and pretext tasks from the natural language processing field to train proteins. For example, modeling protein sequences using LSTM may be a major innovation, but replacing LSTM with Bi-LSTM for stuble performance improvements makes little sense. Now, it is time to step out of this comfort zone of scientific research, and we should no longer be satisfied with simply extending techniques from other domains to the protein domain. PRL is not only a machine learning problem but also a biological problem, so we should consider designing more protein-specific architectures and pretext tasks by incorporating protein-related domain knowledge. In particular, most of the existing work on PRL is based on unimodal protein sequences or structures, and it requires more work exploring sequence-structure co-modeling to fully explore the correspondence between 1D sequences and 3D structures.

\subsubsection{Direction 4: Margin from pre-training to fine-tuning}
Currently, tremendous efforts are focusing on protein pre-training strategies. However, how to fine-tune these pre-trained models to specific downstream tasks is still under-explored. Though numerous strategies have been proposed to address this problem in the fields of computer vision and natural language processing \cite{zhuang2020comprehensive}, they are difficult to be directly applied to proteins. One obstacle to knowledge transfer is the huge variability between different protein datasets, both in terms of sequence length and structural complexity. The second
one is poor generalization of pre-trained models especially for various tasks where collecting labeled data is laborious. Therefore, it is an important issue to design protein-specific techniques to minimize the margin between pre-training and downstream tasks.

\subsubsection{Direction 5: Lack of explainability}
While existing protein representation learning methods have achieved promising results on a variety of downstream tasks, we still know little about what the model has learned from protein data. Which of the feature patterns, sequence fragments, or sequence-structure relationships has been learned? These are important issues for understanding and interpreting model behavior, especially for those privacy-secure tasks such as drug design, but are missing in current PRL works. Overall, the interpretability of PRL methods remains to be explored further in many respects, which helps us understand how the model works and provides a guide for better usage.

\section{Conclusions}
A comprehensive survey of the literature on protein representation learning is conducted in this paper. We develop a general unified framework for PRL methods. Moreover, we systematically divide existing PRL methods into three main categories: sequence-based, structure-based, and sequence-structure co-modeling from three different perspectives, including model architectures, pretext tasks, and downstream applications. Finally, we point out the technical limitations of the current research and provide promising directions for future work on PRL. We hope this survey to pave the way for follow-up AI researchers with no bioinformatics background, setting the stage for the development of more future works.

\bibliographystyle{named}
\bibliography{ijcai22}

\end{document}

%% file: math_commands.tex
\usepackage{amsmath,amsfonts,bm}

\def\eqref#1{equation~\ref{#1}}

\def\1{\bm{1}}

\DeclareMathAlphabet{\mathsfit}{\encodingdefault}{\sfdefault}{m}{sl}
\SetMathAlphabet{\mathsfit}{bold}{\encodingdefault}{\sfdefault}{bx}{n}



%% file: overview.tex
\tikzstyle{leaf}=[
draw=hiddendraw,
rounded corners,
minimum height=1.2em,
fill=hidden-orange!40,
text opacity=1, 
align=center,
fill opacity=.5,  
text=black,
align=left,
font=\scriptsize,
inner xsep=3pt,
inner ysep=1pt,
]

\tikzstyle{level0}=[
draw=hiddendraw,
rounded corners,
minimum height=1.2em,
fill=hidden-cy!20,
text opacity=1, 
align=center,
fill opacity=.5,  
text=black,
align=left,
font=\scriptsize,
inner xsep=3pt,
inner ysep=1pt,
]

\tikzstyle{level1}=[
draw=hiddendraw,
rounded corners,
minimum height=1.2em,
fill=hidden-blue!150,
text opacity=1, 
align=center,
fill opacity=.5,  
text=black,
align=left,
font=\scriptsize,
inner xsep=3pt,
inner ysep=1pt,
]

\tikzstyle{level2}=[
draw=hiddendraw,
rounded corners,
minimum height=1.2em,
fill=output-purple,
text opacity=1, 
align=center,
fill opacity=.5,  
text=black,
align=left,
font=\scriptsize,
inner xsep=3pt,
inner ysep=1pt,
]

\tikzstyle{level3}=[
draw=hiddendraw,
rounded corners,
minimum height=1.2em,
fill=output-green!40,
text opacity=1, 
align=center,
fill opacity=.5,  
text=black,
align=left,
font=\scriptsize,
inner xsep=3pt,
inner ysep=1pt,
]

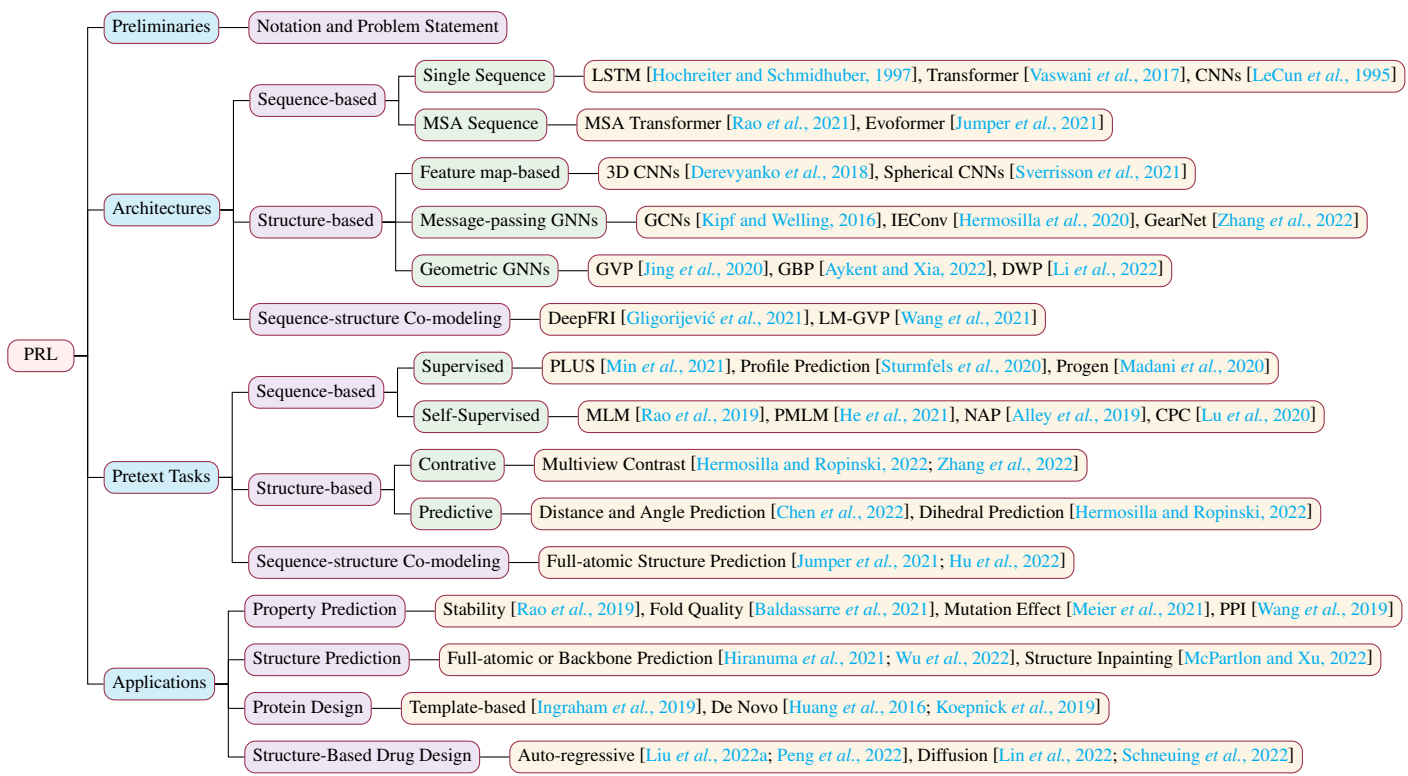
\begin{figure*}[t]
\centering
\begin{forest}
  for tree={
  forked edges,
  grow=east,
  reversed=true,
  anchor=base west,
  parent anchor=east,
  child anchor=west,
  base=middle,
  font=\scriptsize,
  rectangle,
  draw=hiddendraw,
  rounded corners,
  align=left,
  minimum width=2.5em,
  minimum height=1.2em,
    s sep=6pt,
    inner xsep=3pt,
    inner ysep=1pt,
  },
  where level=1{font=\scriptsize}{},
  where level=2{font=\scriptsize}{},
  where level=3{font=\scriptsize}{},
  where level=4{font=\scriptsize}{},
  where level=5{font=\scriptsize}{},
  [PRL, level0
    [Preliminaries,level1
        [Notation and Problem Statement,level2]
    ]
    [Architectures,level1
        [Sequence-based,level2
            [Single Sequence,level3
                [LSTM \cite{hochreiter1997long}{,}~Transformer \cite{vaswani2017attention}{,}~CNNs \cite{lecun1995convolutional},leaf]
            ]
            [MSA Sequence,level3
                [MSA Transformer \cite{rao2021msa}{,}~Evoformer \cite{jumper2021highly},leaf]
            ]
        ]  
        [Structure-based,level2
            [Feature map-based,level3
                [3D CNNs \cite{derevyanko2018deep}{,}~Spherical CNNs \cite{sverrisson2021fast},leaf]
            ]
            [Message-passing GNNs,level3
                [GCNs \cite{kipf2016semi}{,}~IEConv \cite{hermosilla2020intrinsic}{,}~GearNet \cite{zhang2022protein},leaf]
            ]
            [Geometric GNNs,level3
                [GVP \cite{jing2020learning}{,}~GBP \cite{aykent2022gbpnet}{,}~DWP \cite{li2022directed},leaf]
            ]
        ]  
        [Sequence-structure Co-modeling,level2
            [DeepFRI \cite{gligorijevic2021structure}{,}~LM-GVP \cite{wang2021lm},leaf]
        ]  
    ]
    [Pretext Tasks,level1
        [Sequence-based,level2
            [Supervised,level3
                [PLUS \cite{min2021pre}{,}~Profile Prediction \cite{sturmfels2020profile}{,}~Progen \cite{madani2020progen},leaf]
            ]
            [Self-Supervised,level3
                [MLM \cite{rao2019evaluating}{,}~PMLM \cite{he2021pre}{,}~NAP \cite{alley2019unified}{,}~CPC \cite{lu2020self},leaf]
            ]
        ]  
        [Structure-based,level2
            [Contrative,level3
                [Multiview Contrast \cite{hermosilla2022contrastive,zhang2022protein},leaf]
            ]
            [Predictive,level3
                [Distance and Angle Prediction \cite{chen2022structure}{,}~Dihedral Prediction \cite{hermosilla2022contrastive},leaf]
            ]
        ]  
        [Sequence-structure Co-modeling,level2
            [Full-atomic Structure Prediction \cite{jumper2021highly,hu2022exploring},leaf]
        ]  
    ]
    [Applications,level1
        [Property Prediction,level2
            [Stability \cite{rao2019evaluating}{,}~Fold Quality \cite{baldassarre2021graphqa}{,}~Mutation Effect \cite{meier2021language}{,}~PPI \cite{wang2019high},leaf]
        ]  
        [Structure Prediction,level2
            [Full-atomic or Backbone Prediction \cite{hiranuma2021improved,wu2022high}{,}~Structure Inpainting \cite{mcpartlon2022attnpacker},leaf]
        ]  
        [Protein Design,level2
            [Template-based \cite{ingraham2019generative}{,}~De Novo \cite{huang2016coming,koepnick2019novo},leaf]
        ]  
        [Structure-Based Drug Design,level2
            [Auto-regressive \cite{liu2022graphbp,peng2022pocket2mol}{,}~Diffusion \cite{lin2022diffbp,schneuing2022structure},leaf]
        ]  
    ]
  ]
\end{forest}
\caption{A high-level overview of this survey with representative examples.}
\label{fig:2}
\end{figure*}

%% file: table_1.tex
\begin{table*}[!tbp]
\begin{center}
\caption{Summary of representative protein representation learning methods.}
\label{tab:1}
\resizebox{1.0\textwidth}{!}{
\begin{tabular}{lcccl}

\toprule
\textbf{Method} & \textbf{Category} & \textbf{Architecture} & \textbf{Pretext Task} & \textbf{Year} \\ \midrule
Bio2Vec-CNN \cite{wang2019high} & \cellcolor{gray!20}Sequence-based & CNN & - & 2019 \\ \hline
TAPE \cite{rao2019evaluating} & \cellcolor{gray!20}Sequence-based & ResNet, LSTM, Transformer & \tabincell{c}{Masked Language Modeling, \\ Next Amino Acid Prediction} & 2019 \\ \hline
UniRep \cite{alley2019unified} & \cellcolor{gray!20}Sequence-based & Multiplicative LSTM & Next Amino Acid Prediction & 2019 \\ \hline
TripletProt \cite{nourani2020tripletprot} & \cellcolor{gray!20}Sequence-based & Siamese Networks & Contrastive Predictive Coding & 2020 \\ \hline
PLP-CNN \cite{shanehsazzadeh2020transfer} & \cellcolor{gray!20}Sequence-based & CNN & - & 2020 \\ \hline
CPCProt \cite{lu2020self} & \cellcolor{gray!20}Sequence-based & GRU, LSTM & Contrastive Predictive Coding & 2020 \\ \hline
MuPIPR \cite{zhou2020mutation} & \cellcolor{gray!20}Sequence-based & GRU, LSTM & Next Amino Acid Prediction & 2020 \\ \hline
ProtTrans \cite{elnaggar2020prottrans} & \cellcolor{gray!20}Sequence-based & Transformer, Bert, XLNet & Masked Language Modeling  & 2020 \\ \hline
DMPfold \cite{kandathil2020deep} & \cellcolor{gray!20}Sequence-based & GRU, ResNet & - & 2020 \\ \hline
Profile Prediction \cite{sturmfels2020profile} & \cellcolor{gray!20}Sequence-based & Transformer & HMM Profile Prediction  & 2020 \\ \hline
PRoBERTa \cite{nambiar2020transforming} & \cellcolor{gray!20}Sequence-based & Transformer & Masked Language Modeling & 2020 \\ \hline
UDSMProt \cite{strodthoff2020udsmprot} & \cellcolor{gray!20}Sequence-based & LSTM & Next Amino Acid Prediction & 2020 \\ \hline
ESM-1b \cite{rives2021biological} & \cellcolor{gray!20}Sequence-based & Transformer & Masked Language Modeling & 2021 \\ \hline
PMLM \cite{he2021pre} & \cellcolor{gray!20}Sequence-based & Transformer & Pairwise Masked Language Modeling & 2021 \\ \hline
MSA Transformer \cite{rao2021msa} & \cellcolor{gray!20}Sequence-based & MSA Transformer & Masked Language Modeling & 2021 \\ \hline
ProteinLM \cite{xiao2021modeling} & \cellcolor{gray!20}Sequence-based & BERT & Masked Language Modeling & 2021 \\ \hline
PLUS \cite{min2021pre} & \cellcolor{gray!20}Sequence-based & Bidirectional RNN & \tabincell{c}{Masked Language Modeling, \\ Same-Family Prediction} & 2021  \\ \hline
Adversarial MLM \cite{mcdermott2021adversarial} & \cellcolor{gray!20}Sequence-based & Transformer & \tabincell{c}{Masked Language Modeling, \\ Adversarial Training} & 2021  \\ \hline
ProteinBERT \cite{brandes2022proteinbert} & \cellcolor{gray!20}Sequence-based & BERT & Masked Language Modeling & 2022 \\ \hline
CARP \cite{yang2022convolutions} & \cellcolor{gray!20}Sequence-based & CNN & Masked Language Modeling & 2022  \\

\midrule \midrule
3DCNN \cite{derevyanko2018deep} & \cellcolor{red!20}Structure-based & 3DCNN & - & 2018 \\ \hline
IEConv \cite{hermosilla2020intrinsic} & \cellcolor{red!20}Structure-based & IEConv & - & 2020 \\ \hline
GVP-GNN \cite{jing2020learning} & \cellcolor{red!20}Structure-based & GVP & - & 2020 \\ \hline
GraphMS \cite{cheng2021graphms} & \cellcolor{red!20}Structure-based & GCN & Multiview Contrast & 2021 \\ \hline
DL-MSFM \cite{gelman2021neural} & \cellcolor{red!20}Structure-based & GCN & - & 2021 \\ \hline
PG-GNN \cite{xia2021geometric} & \cellcolor{red!20}Structure-based & PG-GNN & - & 2021 \\ \hline
CRL \cite{hermosilla2022contrastive} & \cellcolor{red!20}Structure-based & IEConv & Multiview Contrast & 2022 \\ \hline
DW-GNN \cite{li2022directed} & \cellcolor{red!20}Structure-based & DWP & - & 2022 \\ \hline
GBPNet \cite{aykent2022gbpnet} & \cellcolor{red!20}Structure-based & GBP & - & 2022 \\ \hline
GearNet \cite{zhang2022protein} & \cellcolor{red!20}Structure-based & GearNet & \tabincell{c}{Multiview Contrast, \\ Distance and Dihedral Prediction,  \\ Residue Type Prediction} & 2022 \\ \hline
ATOMRefine \cite{wu2022atomic} & \cellcolor{red!20}Structure-based & SE(3) Transformer & - & 2022 \\ \hline
STEPS \cite{chen2022structure} & \cellcolor{red!20}Structure-based & GIN & Distance and Dihedral Prediction & 2022 \\

 \midrule \midrule
GraphCPI \cite{quan2019graphcpi} & \cellcolor{blue!20}Co-Modeling & CNN, GNN &  - & 2019 \\ \hline
MT-LSTM \cite{bepler2019learning} & \cellcolor{blue!20}Co-Modeling & Bidirectional LSTM &  \tabincell{c}{Contact prediction, \\ Pairwise Similarity Prediction} & 2019 \\ \hline
LM-GVP \cite{wang2021lm} & \cellcolor{blue!20}Co-Modeling & Transformer, GVP & - & 2021 \\ \hline
AlphaFold2 \cite{jumper2021highly} & \cellcolor{blue!20}Co-Modeling & Evoformer &  \tabincell{c}{Masked Language Modeling, \\ Full-atomic Structure Prediction} & 2021 \\ \hline
DeepFRI \cite{gligorijevic2021structure} & \cellcolor{blue!20}Co-Modeling & LSTM, GCN & - & 2021 \\ \hline
HJRSS \cite{mansoor2021toward} & \cellcolor{blue!20}Co-Modeling & SE(3) Transformer &  \tabincell{c}{Masked Language Modeling, \\ Graph Completion} & 2021 \\ \hline
GraSR \cite{xia2022fast} & \cellcolor{blue!20}Co-Modeling & LSTM, GCN & Momentum Contrast & 2022 \\ \hline
CPAC \cite{you2022cross} & \cellcolor{blue!20}Co-Modeling & Hierarchical RNN, GAT & \tabincell{c}{Masked Language Modeling, \\ Graph Completion} & 2022 \\ \hline
MIF-ST \cite{yang2022masked} & \cellcolor{blue!20}Co-Modeling & CNN, GNN & Masked Inverse Folding & 2022 \\ \hline
OmegaFold \cite{wu2022high} & \cellcolor{blue!20}Co-Modeling & Geoformer & \tabincell{c}{Masked Language Modeling, \\ Full-atomic Structure Prediction} & 2022 \\
 \bottomrule
 
\end{tabular}} \vspace{-0.6em}
\end{center}
\end{table*}